\documentclass[conference]{ieeeconf}
\usepackage{times}

\IEEEoverridecommandlockouts                              

\usepackage{multicol}
\usepackage[bookmarks=true]{hyperref}
\usepackage{graphicx}
\usepackage{amsfonts}
\usepackage{amsmath}
\usepackage{amssymb}

\usepackage{amsthm}
\usepackage[ruled,vlined]{algorithm2e}
\usepackage{caption,subcaption}
\usepackage{comment}

\newtheorem{theorem}{Theorem}
\theoremstyle{definition}
\newtheorem{observation}{Observation}

\newcommand{\R}{{\mathbb R}}

\begin{document}

\title{\LARGE\bf Optimal-Horizon Model-Predictive Control with Differential Dynamic Programming}

\author{Kyle Stachowicz and Evangelos A. Theodorou
\thanks{The authors are with the Autonomous Control and Decision Systems Laboratory, Georgia Institute of Technology, Atlanta, GA, USA. Email correspondance to: \texttt{kwstach@gatech.edu}}}

\maketitle

\begin{abstract}


We present an algorithm, based on the Differential Dynamic Programming framework, to handle trajectory optimization problems in which the horizon is determined online rather than fixed a priori. This algorithm exhibits exact one-step convergence for linear, quadratic, time-invariant problems and is fast enough for real-time nonlinear model-predictive control. We show derivations for the nonlinear algorithm in the discrete-time case, and apply this algorithm to a variety of nonlinear problems. Finally, we show the efficacy of the optimal-horizon model-predictive control scheme compared to a standard MPC controller, on an obstacle-avoidance problem with planar robots.



\end{abstract}

\IEEEpeerreviewmaketitle

\section{Introduction}
Trajectory optimization provides a powerful numerical approach to robot motion planning in systems with complex dynamics. The approach is applicable to offline planning problems \cite{Betts1998}, and with increased availability of high-powered onboard compute, to online settings. In particular, model-predictive control approaches \cite{Wang2010Fast, tassa2007receding} present a compelling framework for achieving reactive control in complex environments, by iteratively replanning up until a receding horizon at each instant in time. These approaches allow the controller to approximate a infinite-horizon control policy in a computationally tractable manner.

Planning and control tasks often have the property that they should eventually \textit{complete}, rather than running indefinitely. Parking problems, mobile robot navigation, and swing-up problems share this requirement. Additionally, it is desirable to complete the task quickly (balanced against other objective costs), as shorter solutions may allow the robot to move on to another task.

With standard online trajectory optimization approaches, the time horizon is a fixed design parameter with significant impact on the resulting behavior. While some existing methods can bypass this problem --- particularly direct solution methods using powerful general-purpose optimizers --- these are often too slow to use in an online model-predictive control setting, which can require the optimal control to be computed tens to hundreds of times per second.

In these settings, which in general may include nonlinear dynamics and non-quadratic costs, methods based on the differential dynamic programming algorithm and its derivatives (i.e. iLQR \cite{li2004iterative}) have become very popular due to their relative computational efficiency \cite{Murray1984}. These approaches take the problem's temporal structure into account, using dynamic programming to achieve an efficient solution at each instant in time. In addition, they generalize easily to the stochastic case \cite{theodorou2010stochastic} and have been applied to partially-observable problems through use of belief-space planning \cite{vandenBerg2017}. There exist well-known techniques to easily introduce control constraints \cite{tassa2014control} and arbitrary nonlinear state constraints \cite{xie2017constrained, aoyama2020constrained} using Augmented Lagrangian and penalty methods.

In this work we are primarily concerned with solving the following problem for a system with fixed discrete-time dynamics $f$ and cost $\ell$:
\begin{equation} \label{eq:fixed_dt_formulation}
    \min_{T, u}\sum_{t=0}^{T-1}\ell(x_t, u_t) + \Phi(x_T),\; x_{t+1} = f(x_t, u_t)
\end{equation}

This paper is organized as follows:
in Section \ref{sec:lti}, we present an exact one-pass solution for the linear time-invariant problem in continuous and discrete time. Section \ref{sec:discrete} describes an approximate extension of this solution to the general case and uses it to formulate a real-time model-predictive control algorithm. Section \ref{sec:results} applies this algorithm to several problems, including a nonlinear quadrotor and a point-mass navigation problem.

\section{Related Works}
In the formulation described in equation \ref{eq:fixed_dt_formulation}, the number of knot points $T$ (which also uniquely determines the horizon) must be determined by the solver. This does not fit well into standard optimization frameworks, so prior attempts to solve the optimal-horizon trajectory optimization problem have used different formulations.

Several early methods \cite{jacobson1970differential} \cite{sun2015model} handle a free final time horizon in the continuous-time case by computing the value function's expansions with respect to the horizon. However, these methods are unstable without large regularization and converge slowly in practice, and cannot be applied to discrete-time problems. A very common approach is to fix the number of knot points and optimize the timestep $\Delta t$ \cite{vandenBerg2016, howell2019altro}:
\begin{equation} \label{eq:parametrization_formulation}
    \min_{\Delta t, u} \sum_{k=0}^{N-1} \ell(x_k, u_k)\Delta t + \Phi(x_N),\; \Delta x_{k+1} = f(x_k, u_k)\Delta t
\end{equation}
This method is sensitive to the initial choice of $\Delta t$ \cite{demarchi2019bilevel}, and can be slow to converge. Additionally, the modification of the timestep by the solver can allow solutions that exploit discretization error, for example by making $\Delta t$ large and stepping across an obstacle within a single step \cite{howell2019altro}.

The Timed-Elastic Bands formulation \cite{2015teb} uses a general-purpose nonlinear optimizer to attempt to achieve time-optimal point-to-point robot motion. However, unlike implicit methods like DDP, general-purpose NLP solvers are unable to efficiently exploit the temporal structure present in planning problems. Additionally, problems attempting to solve for the exact time-optimal path must rely on heuristics to avoid chattering when nearing the goal.

In our approach, the expansion of the value function is evaluated (cheaply) at the initial conditions over many possible horizons after each backwards sweep to find an estimate of the objective function for each timestep. The iterative nature of DDP allows us to optimize the horizon and the policy jointly rather than in a bilevel fashion as in \cite{demarchi2019bilevel}. This means that we do not need to wait for DDP to completely converge before adjusting the timestep. 

\section{Linear Time-Invariant Case}
\label{sec:lti}
The linear time-invariant case gives important insight into the structure of the optimal-horizon control problem.
Define the objective function $J(u, T)$ as follows:
\[J(u, T) = \sum_{k=0}^{T-1}\ell(x_k, u_k) + \Phi(x_T)\]
Define the optimal value function (also known as ``cost-to-go'') with horizon $T$ at time $t$ as $V^{t:T}(x)$:
\[V^{t:T}(x) = \min_{u}\left[\sum_{k=t}^{T-1} \ell(x_k, u_k) + \Phi(x_T)\right]\]
Subject to initial condition $x_t = x$ and linear dynamics constraints:
\[x_{k+1} = f(x_k, u_k) = Ax_k + Bu_k\]
As well as quadratic costs:
\[\ell(x, u) = \frac 1 2\left[x^\top Q x + u^\top R u\right]\quad\Phi(x) = \frac 1 2 x^\top Q_f x\]
In the LQR case $V^{t:T}$ is quadratic with form given by $V^{t:T}(x) = \frac 1 2 x^\top P^{t:T} x$. The discrete Ricatti equation states that (with $P = P^{t-1:T}$ and $P' = P^{t:T}$ for clarity):
\[P = A^\top P' A - A^\top P' B (R + B^\top P' B)^{-1} B^\top P' A + Q\]
Thus for a fixed horizon $T$ we can calculate $P^{t-1:T}$ backwards starting at $t = T, P^{T:T} = Q_f$.

\begin{observation}
\label{thm:lti_shift_invariance}
Assume we have a system with time-invariant dynamics and cost. Let $V^{t:T}(x)$ be the value function in quadratic form at time $t$ for the problem with horizon $T$; then $V^{t+d:T+d}(x) = V^{t:T}(x)$.
\end{observation}
\begin{proof}
This follows easily from the control problem:
\[\min_u\sum_{t=0}^{T-1}\ell_t(x_t, u_t) + \Phi(x_T)\]
Then, with $x'_{t'} = x_{t'-\tau}$, $u'_{t'} = u_{t'-\tau}$:
\begin{align*}
    \min_u\sum_{t'=\tau}^{T+\tau-1}\ell_{t'-\tau}(x_{t'-\tau}, u_{t'-\tau}) + \Phi(x_{T'-\tau}) \\
    = \min_u\sum_{t'=\tau}^{T+\tau-1}\ell_{t'}(x'_{t'}, u'_{t'}) + \Phi(x'_{T'})
\end{align*}
\end{proof}
Note that this property still holds for arbitrary dynamics and costs, provided that both are stationary. We will assume stationary dynamics and costs throughout the remainder of this paper.

Using Observation \ref{thm:lti_shift_invariance}, we can rewrite our minimization:
\[\min_{\pi, T} J(\pi, T) = \min_T V^{0:T}_0(x_0) = \min_T \frac{1}{2} x_0^\top P^{-T:0} x_0\]
The right-hand side can be directly evaluated for each possible horizon $T$ to find a minimizer. However, it is possible for some problems that there is no minimizer for $T$ and that we can make $J$ arbitrarily small by increasing $T$ arbitrarily. For example:
\[x_{t+1} = \begin{pmatrix}1 & \alpha \\ 0 & 1\end{pmatrix}x_t + \begin{pmatrix}0 \\ \beta\end{pmatrix}u_t\]
\[\ell(x, u) = \frac 1 2 \lVert u \rVert^2, \Phi(x) = \lVert x \rVert^2\]
In this case $J$ does not achieve its lower-bound $\inf J$ (given by the solution to the infinite-horizon problem) and so $J^*$ does not exist.

We can solve this problem in two ways. Firstly, we can box-constrain $T$ such that $T^- \le T \le T^+$, in which case we will clearly perform finitely many steps of the backwards sweep. Otherwise, if we can find a lower bound $\delta$ such that $0 < \delta \le \ell(x, u)$ then the problem will always be well-formed with a bound on $T$ given by: \[T \le \frac{1}{\delta} J(T, u) \le \frac{1}{2\delta} x_0^\top P_{0, -T} x_0\]

\begin{figure}
    \centering
\includegraphics[width=\linewidth]{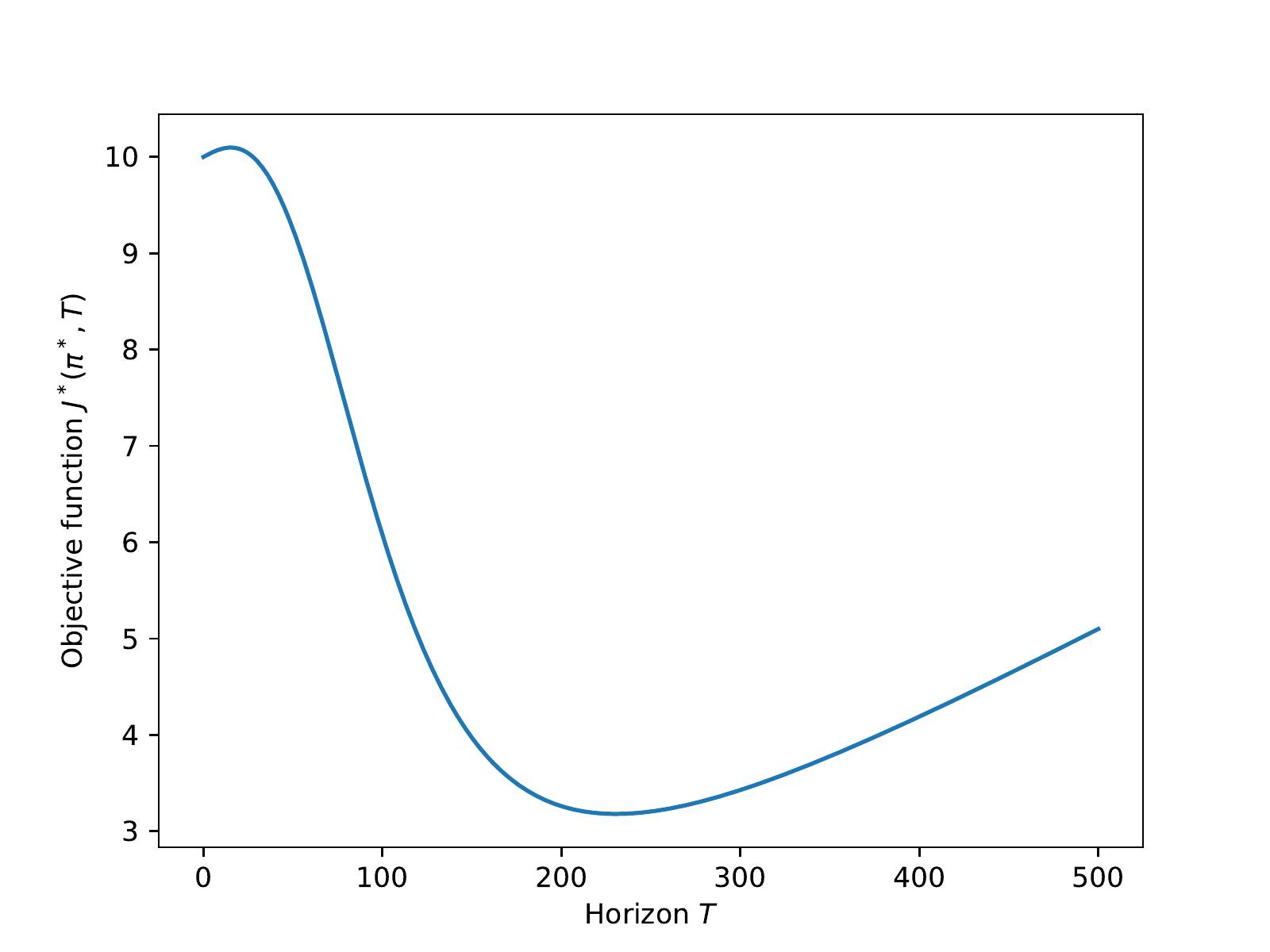}
    \caption{Optimum objective function by horizon $T$ for a LTI system coincides exactly with $\frac{1}{2}x_0^\top P_{\bar T - T} x_0$.}
    \label{fig:lti_objective}
\end{figure}

This second condition can be guaranteed when time is explicitly penalized by adding a constant term to $\ell(x, u)$. This can be achieved in the standard LQR regime using the following state augmentation:
\[\begin{array}{cccc}
    \hat x = \begin{pmatrix}x \\ 1\end{pmatrix} & \hat A = \begin{pmatrix}A & 0 \\ 0 & 1\end{pmatrix} & \hat B = \begin{pmatrix}B \\ 0\end{pmatrix} & \hat Q = \begin{pmatrix}Q & 0 \\ 0 & c_t\end{pmatrix}
\end{array}\]
For any positive $c_t$, this formulation guarantees that the problem will be well-formed assuming $Q$ and $R$ are positive, and the objective function is equivalent to:
\[J(T, u) = \sum_{k=0}^{T-1}\frac 1 2 \left[x_k^\top Q x_k + u_k^\top R u_k\right] + \frac 1 2 x_T^\top Q_f x_T + c_tT\]
Figure \ref{fig:lti_objective} shows a plot of the optimum objective $V^{0:T}(x_0)$ for a linear time-invariant system with positive time penalty $c_t$ and large terminal cost $Q_f$. Notice that it starts large and decreases rapidly as it becomes feasible to reach closer to the goal within the time horizon, before increasing as the $c_t$ term becomes increasingly dominant.



\section{Iterative Algorithm for Nonlinear/Non-Quadratic Case}
\label{sec:discrete}

\subsection{Backwards Sweep}
Here, we will use the notation $V^{t:T}(x)$ to denote the value function at time $t$ with horizon $T$. However, we will drop the horizon (i.e. $V^t(x)$) for clarity when only a single time horizon is under consideration, including in the backwards sweep. This section is standard in the DDP/iLQR literature; for a more thorough treatment refer to \cite{li2004iterative} or \cite{jacobson1970differential}.

We expand the value function around a nominal trajectory $\bar x_t, \bar u_t$, with $\delta x_t = x_t - \bar x_t, \delta u_t = u_t - \bar u_t$.
\begin{equation} \label{eq:value_discrete}
    V^{t}(x) \approx \tilde V^t(x) = \frac{1}{2} \delta x_t^\top V^{t}_{xx} \delta x + V^{t}_{x} \delta x_t + V^{t}_0
\end{equation}
At the final timestep $T$ we simply have:
\[V^{T}(x) = \Phi(x) \approx \frac{1}{2}\delta x^\top\Phi_{xx}\delta x + \Phi_x \delta x + \Phi\]
So $V^{T}_{xx} = \Phi_{xx}$, $V^{T}_{x} = \Phi_x$, and $V^{T} = \Phi(\bar x)$. Then, we can apply Bellman's principle and substitute our value function approximation:
\begin{align} \label{eq:discrete_bellman_expansion}
\begin{split}
    V^{t-1:T}(x) &= \min_u\left[\ell(x, u) + V^{t:T}_{t-1}(f(x, u))\right] \\
    &\approx \frac 1 2 \min_u \begin{pmatrix}\delta x \\ \delta u \\ 1\end{pmatrix}^\top\begin{pmatrix}Q_{xx} & Q_{ux}^\top & Q_x^\top \\ Q_{ux} & Q_{uu} & Q_u^\top \\ Q_x & Q_u & Q\end{pmatrix}\begin{pmatrix}\delta x \\ \delta u \\ 1\end{pmatrix} \\
Q_{xx} &= \ell_{xx} + f_x^\top V^{t+1}_{xx} f_x + V^{t+1}_{x} \cdot f_{xx} \\
Q_{ux} &= \ell_{ux} + f_u^\top V^{t+1}_{xx} f_x + V^{t+1}_{x} \cdot f_{ux} \\
Q_{uu} &= \ell_{uu} + f_u^\top V^{t+1}_{xx} f_u + V^{t+1}_{x} \cdot f_{uu} \\
Q_x &= \ell_x + f^\top_x V^{t+1}_{x} \\
Q_u &= \ell_u + f^\top_u V^{t+1}_{x} \\
Q &= \ell + V^{t+1}_0
\end{split}
\end{align}
Where derivatives of $f$ and $\ell$ are again taken around the nominal state/control pair $\bar x_t, \bar u_t$. In practice, the second-order dynamics terms $f_{xx}, f_{ux}, f_{uu}$ are large tensors with no effect on the converged solution and are often ignored \cite{li2004iterative}, leading to the iLQR formulation. Let $K_t = -Q_{uu}^{-1}Q_{ux}$ and $k_t = -Q_{uu}^{-1}Q_u$. Then, the minimizing control $u_t$ is given by $\delta u_t = K_t\delta x + k_t$, and we have:
\begin{align*}
V^{t}_{xx} &= Q_{xx} - Q_{ux}^\top Q_{uu}^{-1} Q_{ux} \\
V^{t}_{x} &= Q_x - Q_{ux}^\top Q_{uu}^{-1} Q_{u} \\
V^{t}_0 &= Q - \frac{1}{2}Q_{u}^\top Q_{u}^{-1} Q_{u}
\end{align*}
Given the terminal conditions and the backwards recurrence equation we can then solve for $V^t_{xx}, V^t_{x}, V^t_0$ for all $t < T$.

\subsection{Horizon Selection}
\label{sec:discrete_horizon_selection}

Analogously to the LTI case, we see that the value function approximation $V^{t:\bar T}(x)$ provides an estimate for the true value function $V^{0:\bar T - t}(x)$ for $x$ near $x_t$. The minimizing horizon will by definition be achieved by selecting $T$ such that $V^{0:T}(x_0)$ is minimal. While unlike the LTI case we lack an exact description of $V^{0:T}(x)$, we can approximate it using our quadratic value function approximation $\tilde V$. Then, our problem becomes:
\begin{equation}T = \bar T - \operatorname*{min}_{t} \tilde V^{t:\bar T}(x_0)\end{equation}
This optimization problem simply consists of evaluating finitely many quadratic functions at $x_0$ and selecting the minimum, which is computationally trivial. In practice, our approximation is only valid for $x_0$ near $x_t$, which can be interpreted as a constraint $\lVert x_0 - x_t \rVert < \epsilon$ on valid choices of candidate horizon.

This approximation is only locally valid around a particular trajectory, meaning that to increase the horizon from $\bar T$ we need to expand the value function around a longer nominal trajectory. Assume we want to consider candidate horizons up to $\bar T + S$ for some integer $S$. When $(x_0, u_0)$ is a fixed point this can be accomplished by extending $x_{-s} = x_0, u_{-s} = u_0$ for $0 < s \le K$. In the general case, it is only necessary to find an arbitrary dynamically feasible trajectory for $0 < s < K$, for example by integrating backwards.

The error of the value function approximation determines the accuracy of the overall algorithm: this selection process may choose a suboptimal horizon depending on the accuracy of the approximation. In particular, if the true optimal horizon $T^*$ has value $V^*(x_0)$ but is approximated as $\tilde V(x_0) = V^*(x_0) + \epsilon$, then any horizon with value less than $\tilde V(x_0)$ may be chosen instead, giving a suboptimal cost by $\epsilon$. In the next section, a bound on this error is derived in the case of bounded higher-order derivative terms.


\begin{algorithm}
\SetAlgoLined
\SetKwFor{Case}{case}{}{}%
\While{No cost-reducing solution found}{
    $T^* \gets \bar T, J^* \gets \infty$\;
    \For{$T \gets \bar T - S$ \KwTo $\bar T + S$}{
        $t_0 \gets \bar T - T$\;
        $\delta x_0 \gets x_0 - \bar x_{t_0}$\;
        $J_{T} \gets \frac{1}{2}\delta x_0^\top V^{t_0:\bar T}_{xx} \delta x_0 + V^{t_0:\bar T}_{x} \delta x_0 + V^{t_0:\bar T}_0$\;
        \If{$J_T < J^*$}{
            $J^* \gets J_T$\;
            $T^* \gets T$\;
        }
    }
    $J, \bar x', \bar u' \gets $ Rollout policy:
    
    {\Indp $u_t = K_{t_0+t}\delta x_t + \alpha k_{t_0+t} + \bar u_{t_0+t}$\;}
    \uIf{$J < J_{prev}$}{
        \Return{$T^*, \bar x', \bar u'$}\;
    }\uElse{
        Adjust $S$, $\alpha$
    }
}
\Return{$T^*$}
\caption{Horizon Selection and Forwards Pass}
\label{algo:horizon_selection}
\end{algorithm}

\subsection{Analysis of Approximation Error}
\label{sec:analysis}
Throughout this section, for a linear map $f: \R^j \to \R^k$ we denote its operator norm as $\lVert f \rVert$. Additionally, we will use the notation $\lVert G \rVert$ for a $n$-form $G: \R^k \to \R$ to denote the maximum value taken by $G$ on any unit vector, and note that $\lVert G(x) \rVert < \lVert G \rVert \lVert x \rVert^n$.
\begin{theorem}
Let $V^t(x)$ be the true value function for a fixed horizon, and let $\tilde V^t(x)$ be its second-order Taylor approximation around a nominal trajectory $\bar x_t, \bar u_t$. Assume the true optimal policy is given by $\pi$, and that the closed-loop error dynamics $\delta x_{t+1} = f(\bar x_t + \delta x_t, \pi(\bar x_t + \delta x_t))$ are stable at each timestep with all eigenvalues less than some constant $K < 1$. Additionally, assume that the high-order terms are bounded as follows:
\[\lVert \ell_{hhh} \rVert + \lVert V^t_x f_{hhh} \rVert + 3\lVert f_h^\top V^t_{xx}f_{hh}\rVert \le P\]
Where a subscript $h$ denotes the total derivative with respect to $h$ of a function evaluated at $(\bar x + hv, \pi(\bar x + hv))$.
Then, the total approximation error of the value function at any stage is bounded by:
\[\lvert V^t(x) - \tilde V^t(x) \rvert \le \frac{P}{1 - K^3} \lVert x - \bar x \rVert^3 + \mathcal{O}(\lVert x - \bar x \rVert^4)\]
\end{theorem}
\begin{proof}
Because the value function approximation is a Taylor expansion up to degree 2, we only need to consider third-order and higher terms. Let $c = \lVert x_t - \bar x_t \rVert$:
\begin{align*}
    \lvert V^t(x) - \tilde V^t(x) \rvert &\le \left\lvert V^t_{xxx}(x_t - \bar x_t) + \mathcal{O}(c^4)\right\rvert \\
    &\le \left\lvert \max_{v \in S^{n-1}}V^t_{xxx}(v)c^3 + \mathcal{O}(c^4)\right\rvert
\end{align*}
Then, apply the Bellman equation to the third-order term:
\begin{align*}
    \lVert V^t_{xxx} \rVert = \left\lVert \frac{d^3}{dh^3}\left[\ell + V^{t+1} \circ f\right](\bar x + hv, \pi(\bar x + hv))\right\rVert \\
    \le \lVert \ell_{hhh} \rVert + \lVert V^{t+1}_{x}f_{hhh} \rVert + 3\lVert f_h^\top V^{t+1}_{xx}f_{hh} \rVert+\lVert V^{t+1}_{xxx}\rVert \lVert f_{h} \rVert^3 \\
\end{align*}
This yields the recurrence relation $\lVert V^t_{xxx} \rVert \le P + \lVert V^{t+1}_{xxx} \rVert K^3$, which can be solved to yield an upper bound of $\lVert V^t_{xxx} \rVert \le \frac{P}{1-K^3}$. Substituting into our original bound, we get:
\[\lvert V^t(x) - \tilde V^t(x) \rvert \le \frac{P}{1-K^3}\lVert x_t - \bar x_t\rVert^3 + \mathcal{O}(c^4)\]
\end{proof}

This inequality yields several key insights:
\begin{enumerate}
    \item The error bound is only nonzero in the presence of terms ignored by DDP.
    \item If the optimal fixed-horizon solution is stable, the bound exists for any horizon.
    \item The error is cubic in deviation from the trajectory.
\end{enumerate}

Therefore, assuming stationary dynamics and cost functions, the horizon selection algorithm will select a horizon that is at worst $\epsilon$-suboptimal assuming that DDP has converged to an optimal fixed-horizon solution. Note that in practice convergence is not necessary to select a new horizon, as shown in Figure \ref{fig:cost_to_go_compare}, and we can in fact select a new horizon after every iteration of DDP.

\subsection{Forwards Sweep}

Finally, we use our linear controller defined by $K_t, k_t$ to calculate a new nominal trajectory in the forwards pass, starting with our fixed $x_0$:
\begin{equation}
    u_t = K(x - \bar x) + k + \bar u_t \quad\quad x_{t+1} = f(x, u_t)
\end{equation}

These $x_t, u_t$ then become the $\bar x_t, \bar u_t$ for the next iteration. These two procedures repeat until convergence.

The value function and controller yielded by DDP are only valid near the nominal trajectory. To avoid taking large steps outside of this region of validity, it is common to use line search techniques on the forwards pass to ensure convergence. If the forwards pass does not yield a reduction in cost, it can be discarded and a new candidate solution can be computed by decreasing the deviation $k$ until the cost is reduced. In a similar vein, the value function approximation may be invalid for large $\bar T - T$. We can address this issue by only considering $T$ within a window of $\bar T$, and shrinking this window if cost increases.

\begin{figure}
    \centering
    \includegraphics[width=\linewidth]{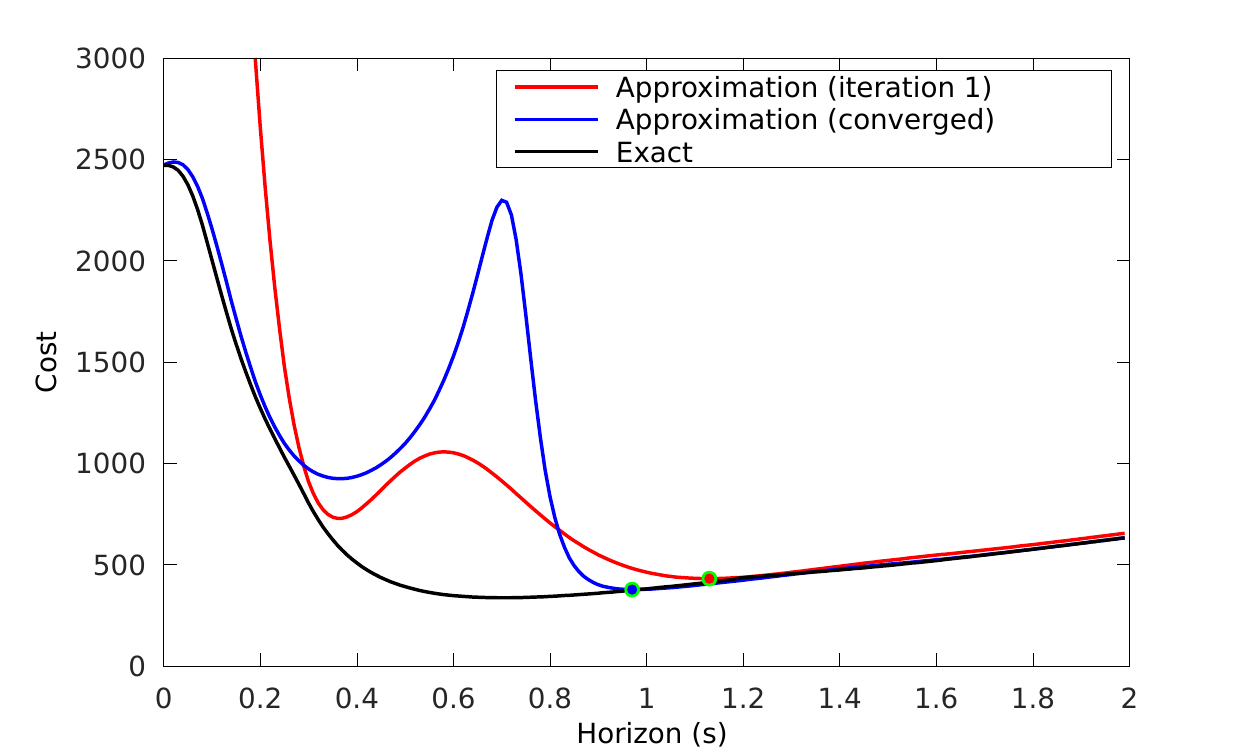}
    \caption{Comparison of $V^{0:T}$ with approximation $V^{t:\bar T}$ after one iteration of DDP and after full convergence. Selected horizons for each case shown with green dots.}
    \label{fig:cost_to_go_compare}
\end{figure}

Figures \ref{fig:cost_to_go_compare} show the value-function approximation after a single iteration of DDP and a fully converged instance of fixed-horizon DDP. While convergence yields a better approximation for the value function, the single-iteration approximation gives a good approximation in a neighborhood of the nominal horizon. This observation indicates that our algorithm is able to operate effectively without the need for a bilevel optimization approach.

\subsection{Model-Predictive Control}
\label{sec:mpc}

Model predictive control is a common approach to handling the dynamically changing environments in which robots often operate. In this scheme, at time $t$, an optimal plan $(x_k, u_k)$ is computed for $t \le k \le t + T$, with $x_t$ as the robot's current state. Then, the first action $u_t$ is applied and a state estimator calculates a new $x_{t+1}$. Finally, a new plan $(x_k, u_k)$ is computed for $t + 1 \le k \le t+T+1$, and the cycle repeats.

Algorithm \ref{alg:mpc} describes a model-predictive scheme for control with optimal-horizon controllers. Unlike the receding-horizon context, the algorithm is also responsible for determining when to terminate the action, which is important for cases in which the robot must balance time-to-completion of the task with accuracy --- for example if it must begin a separate task after completing a motion.

When the system dynamics and cost function are known exactly in advance by the controller, the MPC formulation is unnecessary (the update rule becomes $\pi \gets \pi_{t+1:t+T}$ - in other words, the first timestep is dropped). However, the policy update will be nontrivial in the presence of any of the following disturbances:
\begin{itemize}
    \item Noisy or incorrect dynamics model, such that the real value of $x_{t+1}$ is not identical to its predicted value
    \item Uncertain future cost landscape - for example, randomly moving obstacles
\end{itemize}
In real-world robotic systems, both of these conditions hold and simply tracking a pre-planned trajectory will fail, highlighting the importance of MPC controllers.

\begin{algorithm}
\SetAlgoLined
$(\bar x, \bar u), \bar T \gets $ Initial computed trajectory\;
\While{$\bar T > 1$}{
    $(\bar x, \bar u), \bar T \gets $ OptimizeTrajectory($\bar x, \bar u, \bar T$)\;
    Apply action $\bar u_0$\;
    Drop $(\bar x_0, \bar u_0)$\;
    $\bar T \gets \bar T - 1$\;
}
\caption{Model-Predictive Control}
\label{alg:mpc}
\end{algorithm}

\section{Results}
\label{sec:results}

We compared our algorithm against several existing solution methods for optimal-horizon problems: Sun et. al \cite{sun2015model}, a continuous-time DDP approach, and the standard direct transcription approach using the an off-the-shelf interior-point NLP solver \cite{ipopt}. Table \ref{tab:compute} gives a summary of the three solvers' performance on each of the following four problems.

We implemented the presented nonlinear optimal-horizon trajectory optimization algorithm in C++, with derivatives computed numerically. All tests were performed performed on a desktop computer with a Ryzen 9 3950X processor.

\subsection{Linear System}
We validated the results of Section \ref{sec:lti} by running our algorithm against a simple linear time-invariant double-integrator. As expected, only a single iteration was required to compute an optimal horizon and policy.

\subsection{Cartpole Swing-Up}

The goal of the swing-up task is to cause a pendulum to arrive at the upright position $\theta = \pi$ with nearly zero velocity as quickly as possible, through linear motion of a cart to which its base is attached. The objective function consists of a quadratic running-cost on $\dot x$ and on $\dot \theta$ and time-penalization $c_t$ (a constant term added to $\ell$), as well as a large terminal cost on $(\theta - \pi)^2$ and terminal velocities.

We swept the parameter $c_t$ to find its effect on optimal horizon, as shown in Table \ref{tab:cartpole_opt_horizon}. The ``exact'' horizons in this table are computed exhaustively using a naive brute-force algorithm by applying the fixed-horizon DDP algorithm to each possible horizon. Our algorithm recovered a optimal or indistinguishable-from-optimal horizon in every case except $c_t = 100$ (where it still achieves nearly optimal cost).


\begin{table}[]
    \vspace{10pt}
    \centering
    \begin{tabular}{|c|c|c|c|}
        \hline
        $c_t$ & Ours & Exact & Cost \% error \\
        \hline
        1.0 & 3.31 & 3.31 & 0.00 \\
        3.0 & 2.53 & 2.54 & 0.00 \\
        10.0 & 1.79 & 1.8 & 0.01 \\
        30.0 & 1.65 & 1.65 & 0.00 \\
        100.0 & 1.47 & 1.41 & 0.27 \\
        \hline
    \end{tabular}
    \caption{Optimal horizons for cartpole on discrete-time problem with varying $c_t$}
    \label{tab:cartpole_opt_horizon}
\end{table}


\begin{figure}
    \centering
    \includegraphics[width=\linewidth]{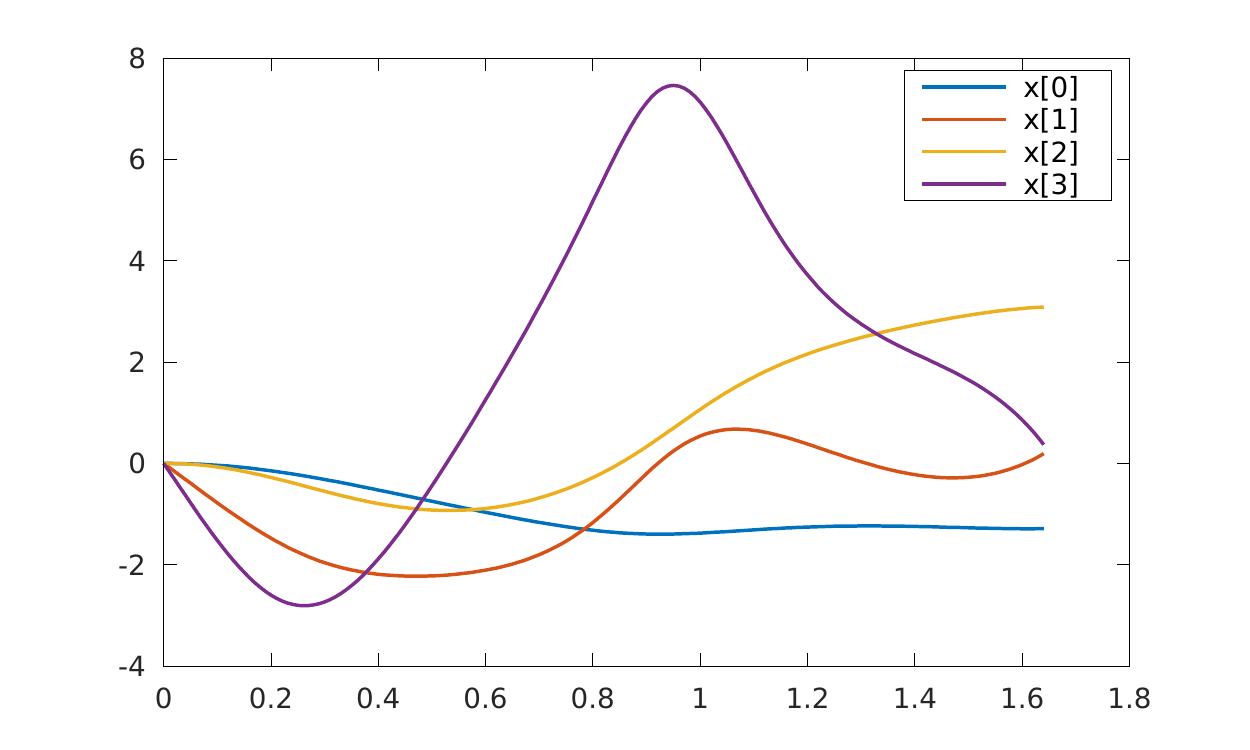}
    \caption{Solution to optimal-horizon cartpole problem with $c_t = 30$}
    \label{fig:cartpole_states}
\end{figure}

\begin{figure}
    \centering
    \includegraphics[width=\linewidth]{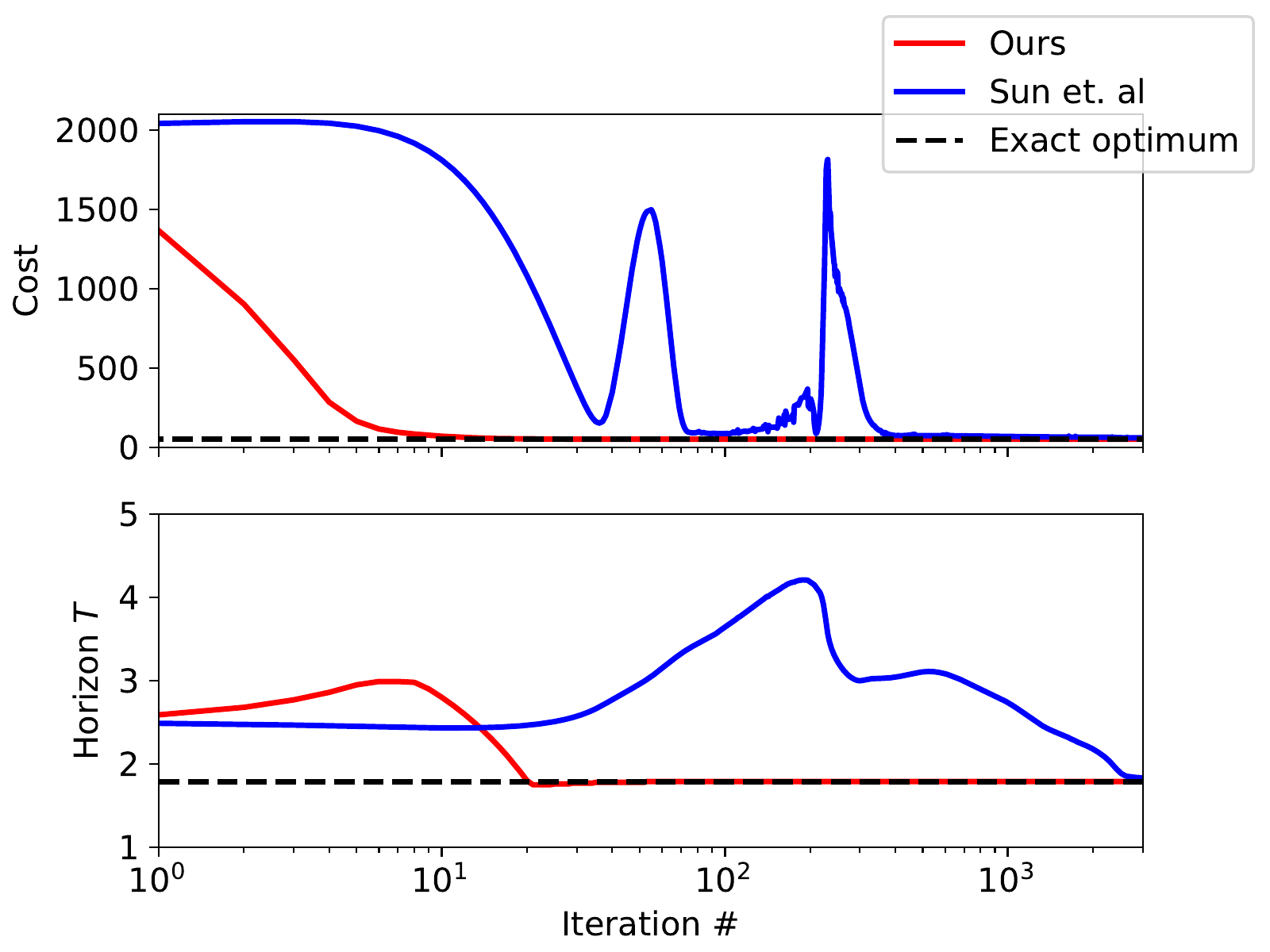}
    \caption{Comparison between progression of optimization between our method and previous work \cite{sun2015model} on the cartpole problem. IPOPT is not shown, as pre-convergence results are infeasible.}
    \label{fig:comparison}
\end{figure}

Figure \ref{fig:cartpole_states} shows one solution for cartpole with $c_t = 30$. This solution took roughly 8.6 milliseconds to compute.
In general, our implementation converges in substantially fewer iterations than both existing DDP-based and general NLP-solver approaches. Figure \ref{fig:comparison} shows the progression of the optimization problem for each algorithm.

\subsection{Quadrotor}
We also applied the algorithm to generate optimal-horizon trajectories for a nonlinear 3D quadrotor model with 12 state dimensions and 4 control dimensions \cite{Sabatino2015QuadrotorCM}. The cost function for this simulation is a simple quadratic function of state and control, with the goal of reaching a set final state. In particular, we quadratically penalize deviation from a desired nominal control (in our case, an upward thrust equal to the force of gravity) and penalize the total time. This problem is numerically challenging and was not solved by the min-time IPOPT formulation or the continuous-time formulation from Sun et. al \cite{sun2015model}.

\begin{figure}
    \centering 
    \begin{subfigure}{0.24\textwidth}
      \includegraphics[trim=70 30 70 30,clip,width=\linewidth]{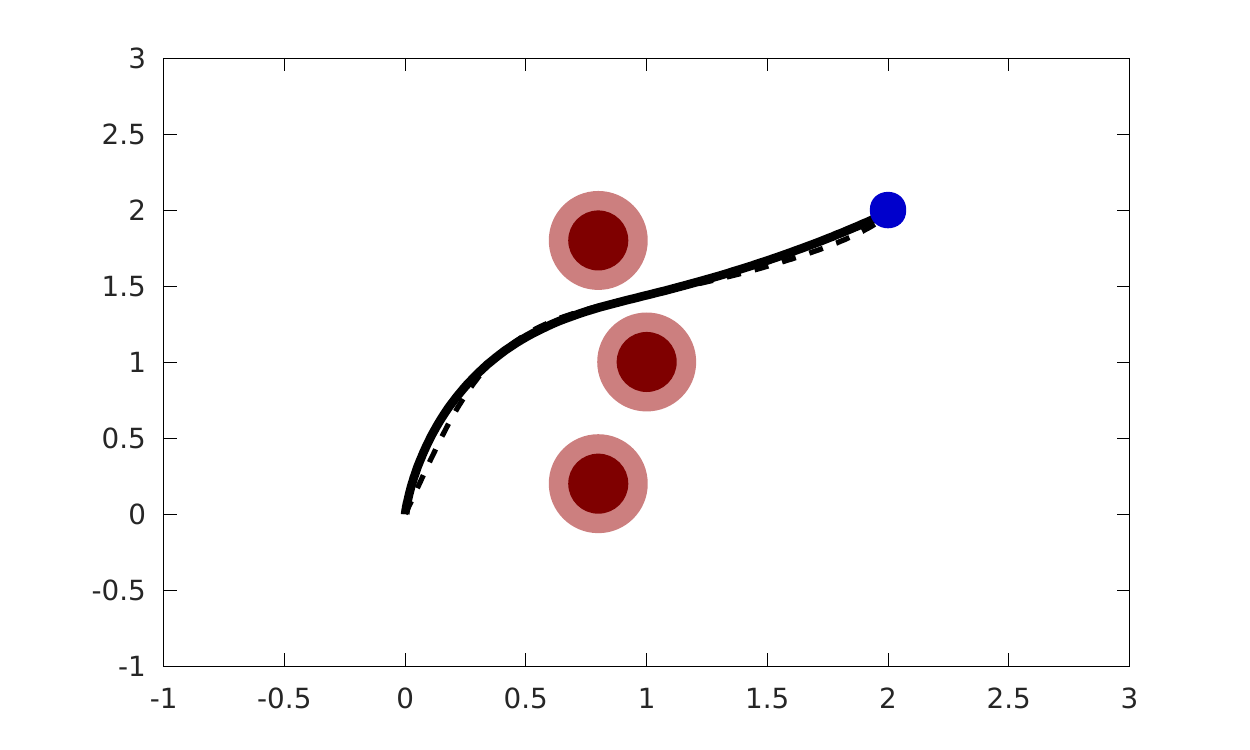}
      \caption{$t = 0$}
    \end{subfigure}\hfil 
    \begin{subfigure}{0.24\textwidth}
      \includegraphics[trim=70 30 70 30,clip,width=\linewidth]{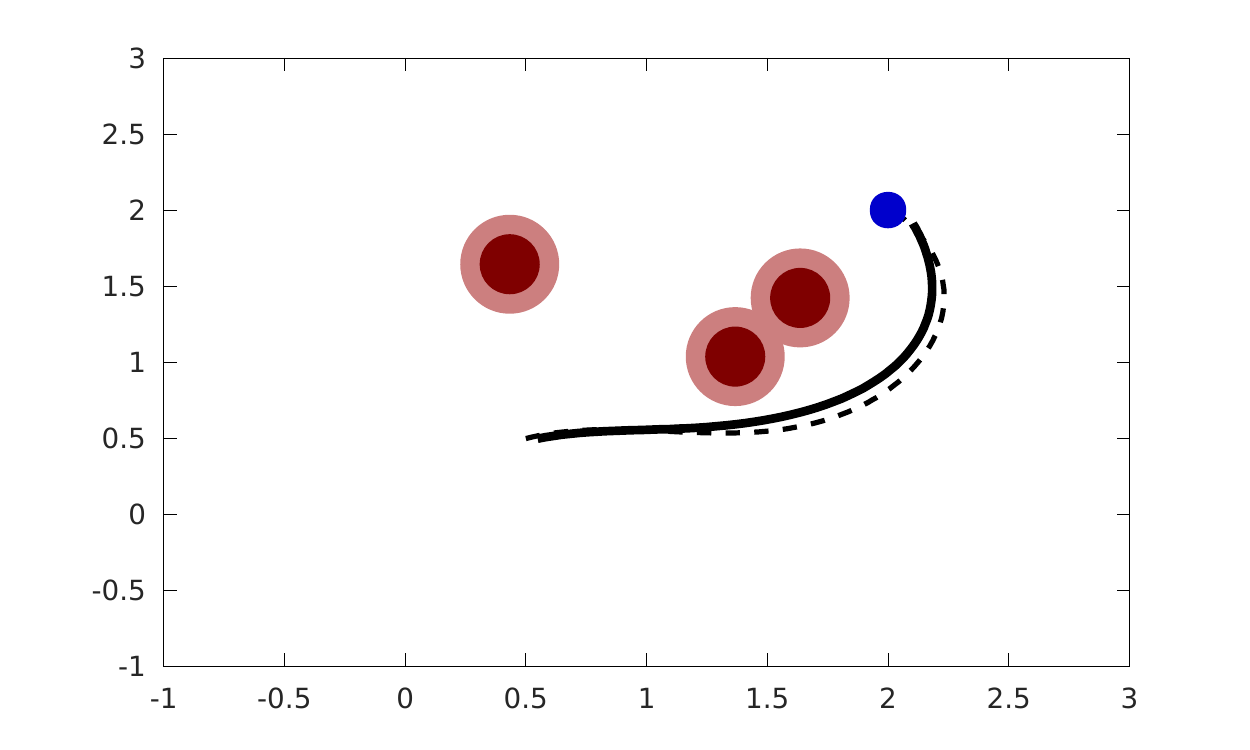}
      \caption{$t = 40$}
    \end{subfigure}
    \begin{subfigure}{0.24\textwidth}
      \includegraphics[trim=70 30 70 30,clip,width=\linewidth]{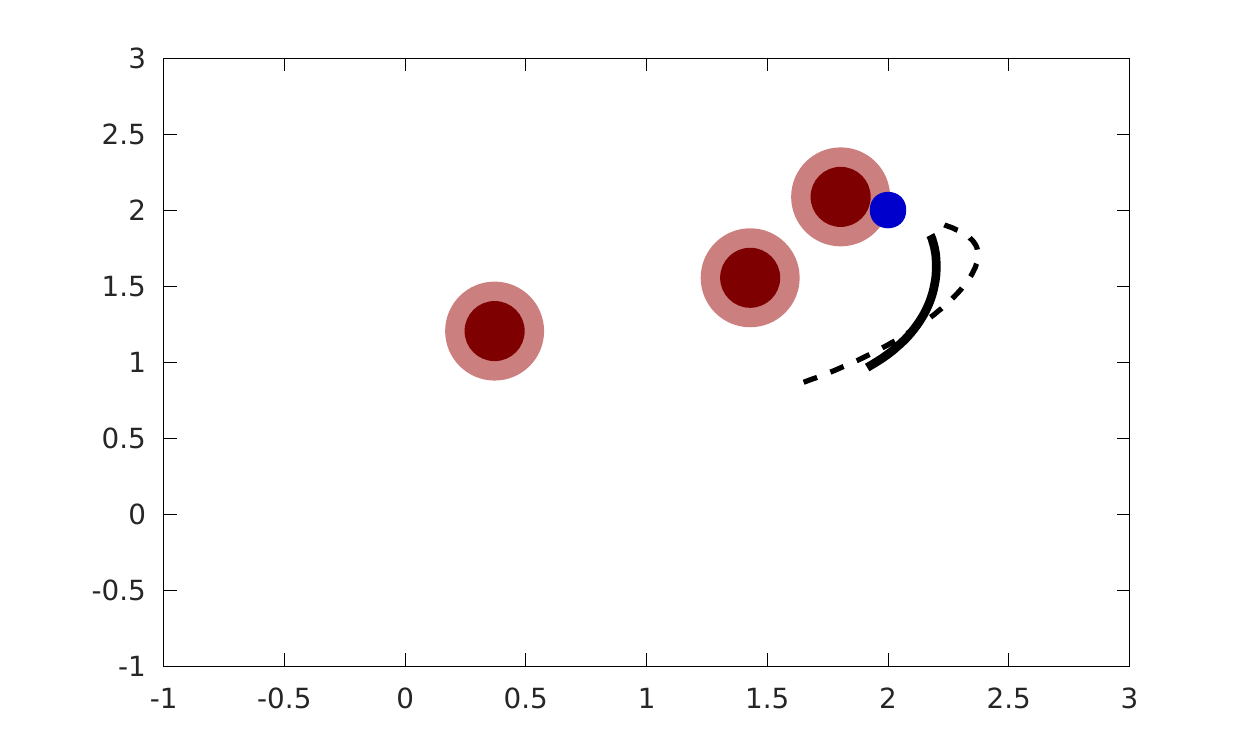}
      \caption{$t = 80$}
    \end{subfigure}\hfil
    \begin{subfigure}{0.24\textwidth}
      \includegraphics[trim=70 30 70 30,clip,width=\linewidth]{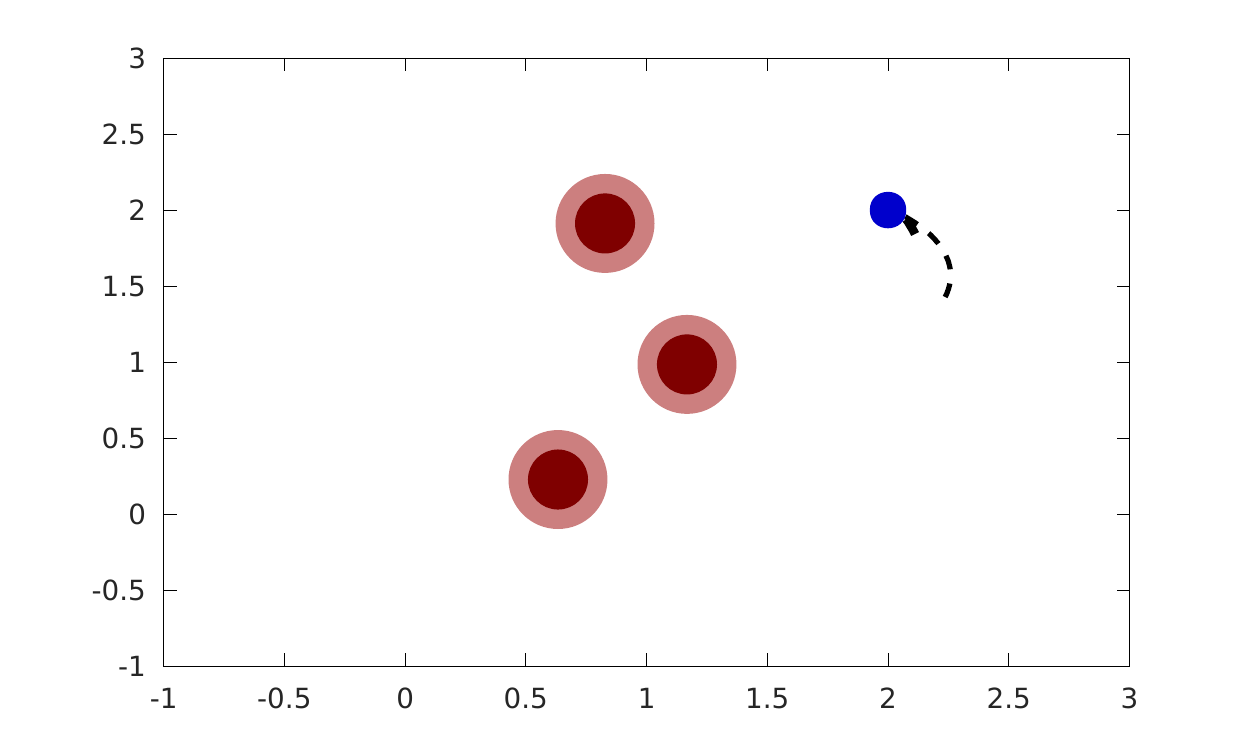}
      \caption{$t = 120$}
    \end{subfigure}
    \caption{Progression of optimal-horizon model-predictive formulation (solid) vs receding-horizon (dashed). The optimal-horizon MPC is able to reach the goal in finite time, while the receding-horizon approach only displays exponential convergence.}
    \label{fig:mpc}
\end{figure}

\subsection{Point-Mass Robot MPC}
Finally, we applied the MPC algorithm described in Section \ref{sec:mpc} to the obstacle-avoidance problem for a point-mass mobile robot. In this problem, the robot must reach a pre-specified goal position while avoiding obstacles whose motion is unknown to the controller. We used a double-integrator dynamics model in which omnidirectional acceleration is directly controlled. Circular obstacles were used with cost $\exp({-\frac{\lVert x - o \rVert^2}{2r^2}})$, where the obstacle has position $o$ and radius $r$, and obstacles were moved in a manner unknown a priori to the planner.
The terminal cost was taken to be a quadratic penalizing deviation from the goal. Again, the $c_t$ term incentivizes the robot to arrive at the goal more quickly.

Figure \ref{fig:mpc} shows four snapshots of a particular experiment with MPC. The planner is able to dynamically adjust the horizon when the obstacle locations change, and ends the episode once it is close to the goal. As the two trajectories approach the goal the receding-horizon implementation only converges exponentially while the optimal-horizon implementation is able to complete in finite time. The MPC step takes on average 5 milliseconds to recompute for each stage when warm-started, easily fast enough for real-time use.


\begin{table}[]
\vspace{10pt}
\centering
\begin{tabular}{l|cc|cc|cc|}
\cline{2-7}
\textbf{}                              & \multicolumn{2}{c|}{\textbf{Ours}} & \multicolumn{2}{c|}{\textbf{Sun et. al}} & \multicolumn{2}{c|}{\textbf{IPOPT}} \\ \hline
\multicolumn{1}{|l|}{\textbf{Problem}} & \textbf{Time}   & \textbf{Iter.}   & \textbf{Time}      & \textbf{Iter.}      & \textbf{Time}    & \textbf{Iter.}   \\ \hline
\multicolumn{1}{|l|}{Linear System}    & 1ms             & 1                & 4.76s              & 100                 & 0.223s           & 37               \\
\multicolumn{1}{|l|}{Cartpole}         & 9ms             & 35               & 213s               & 3000                & 6.415s           & 1223             \\
\multicolumn{1}{|l|}{Quadrotor}        & 101ms           & 25               & \multicolumn{2}{c|}{---}                 & \multicolumn{2}{c|}{---}            \\
\multicolumn{1}{|l|}{Navigation}       & 15ms            & 26               & \multicolumn{2}{c|}{---}                 & 10.273s          & 507              \\
\multicolumn{1}{|l|}{Nav. (MPC)}       & 5ms             & 9                & \multicolumn{2}{c|}{---}                 & 0.581s           & 31               \\ \hline
\end{tabular}
\caption{Comparison of computation times and number of iterations between our algorithm, the continuous-time solver presented in Sun et. al \cite{sun2015model}, and an off-the-shelf optimizer \cite{ipopt} with problem formulation as described in Equation \ref{eq:parametrization_formulation}. Entries marked with a dash were unable to find a solution.}
\label{tab:compute}
\end{table}



\section{Conclusion}
\label{sec:conclusion}
The problem of optimal-horizon planning and control is fundamental in robotic systems and is motivated by common practical problems in the robotics literature. By explicitly including the time horizon in the optimization problem, it is possible to more naturally represent problems in robotics often approximated by infinite-horizon controllers or approximations thereof.
Our algorithm is able to generate solutions to the optimal-horizon control problem then either of the existing approaches tested.

Another key application of the optimal-horizon MPC problem in robotics comes in the form of explicitly handling stochastic dynamics. By extending this framework to stochastic problems, it will be possible to tackle a much wider array of real-world problems. Additionally, it may be possible to extend a similar approach to derive optimal-horizon variants of other commonly-used MPC solvers such as model-predictive path integral \cite{williams2017information} for use in model-based reinforcement learning.

\pagebreak
\bibliographystyle{IEEEtran}
\bibliography{references}


\end{document}